%% file: root.tex
\documentclass[letterpaper,10pt,conference]{ieeeconf}

\IEEEoverridecommandlockouts
\usepackage{amsmath}
\usepackage{amssymb}
\usepackage{algorithm}
\usepackage[noEnd=true,indLines=true]{algpseudocodex}
\usepackage{booktabs}
\usepackage{capt-of}
\usepackage{cite}
\usepackage{graphicx}
\usepackage{multirow}
\usepackage{placeins}
\usepackage{xspace}
\usepackage[hidelinks]{hyperref}

\newif\ificrasubmission
\icrasubmissionfalse

\title{\LARGE \bf
PEARS: Physical-Prior-Guided Efficient Adaptation via \\Failure Reasoning and Diffusion Steering for Tactile Manipulation
}

\newcommand{\methodname}{PEARS\xspace}

\ificrasubmission
  \author{Authors}
  \ifdefined
  \fi
\else
  \author{Kun Song$^{1,*}$, Yiming Wang$^{2,3,*}$, Yilin Chen$^{4,*}$, Tianyi Ding$^{2,*}$,\\
    Jiaxin Tian$^{2}$, Tianqi Gong$^{2}$, Daolin Ma$^{2,3,\dagger}$, Jia Pan$^{1,\dagger}$%
    \thanks{This work was supported by the National Natural Science Foundation of China (NSFC) under Grant Nos. 12522202 and 12272220.}%
    \thanks{$^{1}$The University of Hong Kong;
      $^{2}$Shanghai Jiao Tong University;
      $^{3}$Xense Robotics;
      $^{4}$University of Pennsylvania.}%
    \thanks{$^{*}$These authors contributed equally to this work.
      $^{\dagger}$Co-corresponding authors: Daolin Ma and Jia Pan.}%
  }
\fi

\IEEEaftertitletext{%
  \begin{minipage}{\textwidth}
    \centering
    \includegraphics[width=\textwidth]{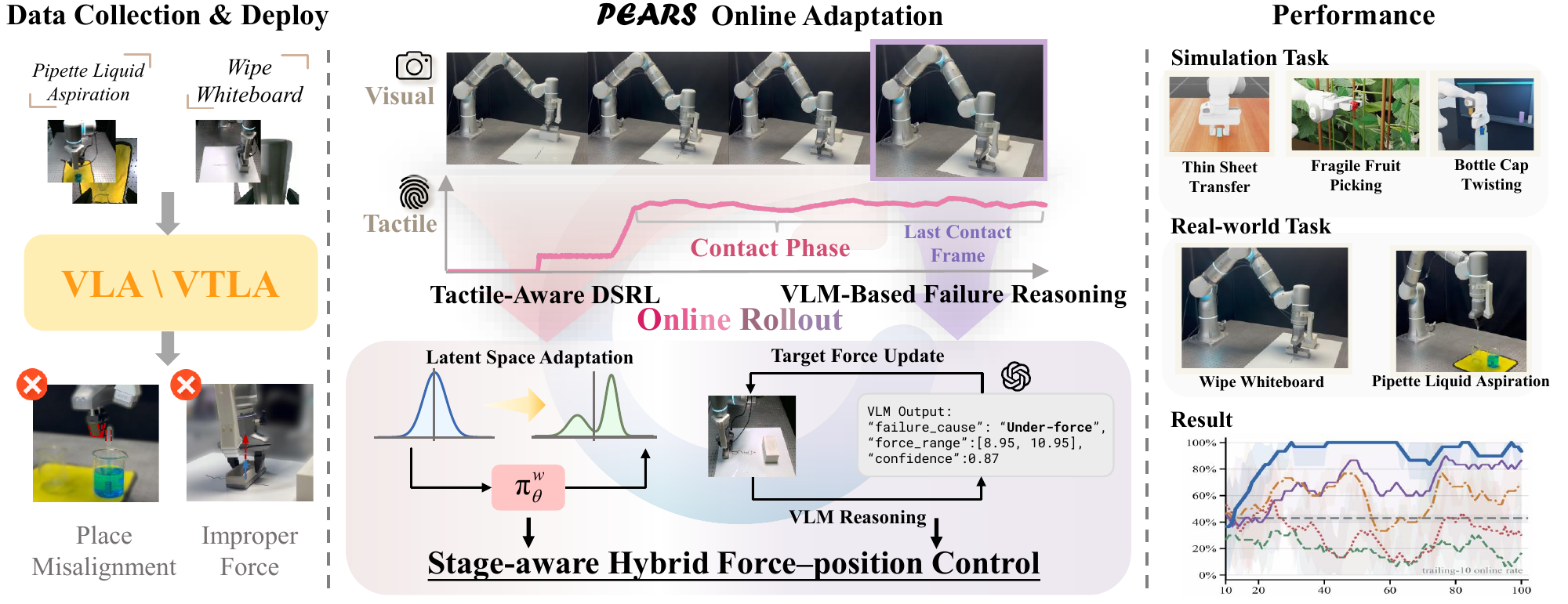}
    \captionof{figure}{Overview of \methodname. Tactile-aware DSRL and
    the physics-guided force reasoning (PFR) module jointly accelerate online adaptation.}
    \label{fig:rapid_tactile_overview}
  \end{minipage}%
  \vspace{0.6em}%
}

\begin{document}

\bstctlcite{IEEEexample:BSTcontrol}

\maketitle
\thispagestyle{empty}
\pagestyle{empty}

\begin{abstract}
Pretrained robotic policies can suffer substantial performance degradation
under out-of-distribution (OOD) conditions encountered during deployment, motivating
post-training through real-world interaction. However, reinforcement-learning
(RL)-based post-training typically requires substantial environment interactions, a
burden that is especially significant in manipulation, where each trial
can be slow, costly, or destructive. Therefore, we present \methodname, a
physics-prior-guided hybrid RL framework for sample-efficient online adaptation
of pretrained policies with tactile feedback. After each episode, its
physics-guided force reasoning (PFR) module uses physical priors encoded in a
vision--language model (VLM) to diagnose failures from the visual outcome and
tactile interaction history and update task-appropriate contact-force
bounds. A high-frequency hybrid force--position controller then
enforces these bounds during contact. Complementarily,
tactile-conditioned diffusion steering reinforcement learning adjusts the
latent noise of the frozen flow-matching policy to correct errors in
free-space motion and contact timing without updating the base model.
In simulation, \methodname improves success rates by 12.4--37.4 percentage points over the strongest per-task baselines.
\methodname also reduces the number of interaction episodes required for a certain success threshold by up to 53.2\% relative to the fastest baseline.
In real-world experiments, \methodname achieves success
rates of 95\% on Whiteboard Erasing and 90\% on Pipette Liquid Aspiration.
These
results show that combining the PFR module with policy
steering can accelerate adaptation while reducing costly interactions.
The project website is available at \url{https://song-kun.github.io/pears}.
\end{abstract}

\input{sections/introduction}
\input{sections/related_work}
\input{sections/method}
\input{sections/exp}
\input{sections/conclusion}

\bibliographystyle{IEEEtran}
\bibliography{references}

\end{document}

%% file: sections/introduction.tex
\section{Introduction}
Large-scale pretrained vision–language–action (VLA) models~\cite{intelligence2025pi_},~\cite{gr00tn1_2025} have recently emerged as an important paradigm for robotic manipulation, achieving remarkable progress in task generalization across diverse real-world environments. For contact-rich tasks, such as surface wiping and grasping fragile objects, vision–tactile–language–action (VTLA) models~\cite{Yuan2026FTP1AG} extend this paradigm by incorporating tactile feedback to sense and regulate physical interactions with the environment. 
However, pretrained VLA and VTLA policies can perform poorly during deployment when scene configurations, object properties, or contact dynamics differ from those encountered during training. Such out-of-distribution (OOD) shifts can lead to positioning errors, mistimed contact, or inappropriate interaction forces~\cite{paresmorlans2026fact}.

Reinforcement learning (RL) post-training with tactile feedback~\cite{Ma2026TacCoRLIT,Yu2026OmniTacTunePR} provides a direct route for adapting pretrained policies to the target deployment environment through autonomous interaction.
However, traditional RL-based algorithms demand substantial interaction with the environment, which can be time-inefficient and high-cost.
For example, a failed online interaction in an industrial operation or chemical laboratory procedure may damage equipment or compromise the experimental setup.
This interaction cost is compounded by conventional online RL algorithms, which often rely on stochastic trial-and-error exploration and must visit a sufficiently diverse set of task-relevant states before effective behaviors can be learned.
Therefore, how to develop an efficient real-world RL fine-tuning method for contact-rich manipulation with tactile feedback is still an open-problem.

To this end, we introduce \textbf{\methodname}, a physics-prior-guided framework for interaction-efficient real-world adaptation in tactile manipulation. Its \textbf{physics-guided force reasoning (PFR)} module uses physical priors encoded in a vision–language model (VLM) to reason about episode-level failures and guide contact-force adaptation, rather than relying solely on random RL exploration.
After each episode, the PFR module jointly interprets the visual outcome and measured interaction forces to diagnose the failure cause and update task-specific contact-force bounds. During contact, a high-frequency hybrid force–position controller enforces these bounds along the force-control axes while maintaining position tracking in the remaining directions. Complementarily, tactile-conditioned DSRL~\cite{Wagenmaker2025SteeringYD} steers the frozen policy through latent-noise adaptation to correct positioning and contact-timing errors without updating the base model. By separating force adaptation through PFR from latent policy steering, \methodname reduces its reliance on costly real-world trial-and-error exploration.

We evaluated \methodname on three simulated tasks and two real-world manipulation tasks. Experimental results demonstrate that \methodname achieves superior performance, significantly outperforming representative policy adaptation baselines.
Our contributions are threefold:
\begin{itemize}
    \item We present \textbf{\methodname}, an interaction-efficient online fine-tuning framework that combines noise-space steering with VLM-based force reasoning using tactile sensing.
    \item We develop the \textbf{PFR module}, which uses physical priors encoded in VLMs to diagnose failure causes and adapt task-specific contact-force bounds, reducing interaction numbers required for policy fine-tuning.
    \item We conduct extensive experiments across simulated and real-world contact-rich manipulation tasks, demonstrating the effectiveness and efficiency of \methodname.
\end{itemize}

%% file: sections/related_work.tex
\section{Related Work}

\subsection{Tactile-Aware Visuomotor Policies}

Recent studies have increasingly incorporated tactile sensing into visuomotor policies.
These approaches employ tactile encoders~\cite{Zhao2025PolyTouchAR},~\cite{Higuera2024SparshST},~\cite{Chen2026UniVTACAU},~\cite{Xu2026VTMUSEMU} cross-modal alignment~\cite{Jones2025BeyondSF},~\cite{zhang2026touchguide},~\cite{Huang2026TaFVLATA} 
or auxiliary tactile supervision~\cite{He2025ViTacFormerLC} to leverage learned tactile representations to 
guide action generation.~\cite{Zheng2026OmniVTAVW},~\cite{n0vtla2026},~\cite{Yuan2026FTP1AG},~\cite{Niu2026TRexTD} further pretrain vision–tactile–language–action (VTLA) models on tactile datasets collected across diverse tasks and heterogeneous tactile sensor configurations, enabling VTLA models to learn more unified and transferable tactile representations.

\subsection{Reactive Tactile Policies for Contact-Rich Manipulation}

Contact-rich manipulation requires high-frequency responses to tactile feedback. FOAR~\cite{he2024force}
uses predicted contact states to dynamically modulate vision-force fusion and trigger local action corrections.
RDP~\cite{xue2025reactivedp} introduces a slow-fast architecture that pairs slow action chunk generation with 
a fast tactile-conditioned decoder, enabling within-chunk corrections. ForcePolicy~\cite{fang2026forcepolicy}
and ForceVLA2~\cite{Li2026ForceVLA2UH} employ force feedback to perform high-frequency hybrid force-position 
control. This reactive design has been extended to tactile-aware VLA models. VLA-Touch~\cite{Bi2025VLATouchEV} 
uses tactile feedback to support both semantic replanning and action refinement. AT-VLA~\cite{Li2026ATVLAAT} 
separates slow vision-language reasoning from fast tactile control. FAVLA~\cite{Li2026FAVLAAF} adaptively adjusts 
the action expert invocation frequency based on predicted force variations. UniTacVLA~\cite{Zhang2026UniTacVLAUT} combines predicted and measured tactile signals to apply high-frequency corrections to low-frequency action chunks. 

\subsection{Reinforcement Learning for Tactile Policy Adaptation}

Beyond behavior cloning (BC), RL can further improve tactile policies through autonomous interaction with the environment. VLA-RL~\cite{Lu2025VLARLTM} optimizes autoregressive VLAs through trajectory-level RL. SimpleVLA-RL~\cite{Li2025SimpleVLARLSV} presents an efficient and scalable RL framework tailored to VLAs. $\pi^{*}_{0.6}$ proposes RECAP~\cite{Intelligence202506AV}, which trains a value function on demonstrations, autonomous rollouts, and expert interventions.
Rather than directly updating VLA parameters, DSRL~\cite{Wagenmaker2025SteeringYD} treats the initial noise of a diffusion or flow-matching policy as the RL latent action space and learns a noise policy, enabling effective autonomous policy improvement in the real world. 
RL Token~\cite{Xu2026RLTB} compresses task-relevant internal VLA representations into a compact RL token and trains a lightweight actor-critic head on this representation to refine actions while anchoring behavior to the base VLA, enabling sample-efficient online adaptation on real robots.
Recent work extends RL-based VLA adaptation to contact-rich manipulation through tactile feedback. TacCoRL~\cite{Ma2026TacCoRLIT} improves tactile-conditioned VLA policies through simulation-based RL. OmniTacTune~\cite{Yu2026OmniTacTunePR} learns a tactile residual policy through online RL to provide contact-aware action corrections. TORL-VLA~\cite{Zheng2026TORLVLATG} freezes a pretrained tactile-aware VLA during online RL and employs an intervention-censored critic for real-world adaptation.
However, most existing RL approaches require substantial environment interaction for stochastic exploration and lack explicit directional guidance, limiting their sample efficiency. To address these limitations, we propose the \methodname framework.

%% file: sections/method.tex
\section{Preliminaries and Problem Formulation}
\subsection{Flow Matching Policy for Imitation Learning}

Behavior cloning (BC) learns a policy from a demonstration dataset
$\mathcal{D}=\{(\mathbf{s}_t,\mathbf{a}_t)\}$, where $s_t \in \mathcal{S}$ is the state and
$\mathbf{a}_t$ is the demonstrated action chunk. For a
vision--language--action (VLA) policy, $\mathbf{s}_t$ contains visual observations,
proprioception, and the language instruction~\cite{black2024pi0}; a
vision--tactile--language--action (VTLA) policy additionally includes tactile
observations~\cite{Yuan2026FTP1AG}. 
In both cases, diffusion and flow-matching policies learn the conditional action distribution by mapping a noise $\boldsymbol{w}$ to the demonstrated action distribution. 
$
    \pi_{\mathrm{BC}}: \mathcal{S}\times \mathcal{W} \rightarrow \mathcal{A}.
$
At inference time, the policy first samples an initial noise action
and then transforms it into a clean action
chunk through an iterative sampling process~\cite{Wagenmaker2025SteeringYD}.

\subsection{Diffusion Steering Reinforcement Learning}

A Markov decision process (MDP) can be denoted using
$
    \mathcal{M}=(\mathcal{S},\mathcal{A},P,R,\gamma),
$
where $\mathbf{s}_t\in\mathcal{S}$ is the state,
$\mathbf{a}_t\in\mathcal{A}$ is the physical action, and $P$, $R$, and
$\gamma$ denote the transition function, reward function, and discount factor,
respectively.

Following DSRL~\cite{Wagenmaker2025SteeringYD}, we do not optimize the physical
action directly. Instead, the RL policy selects the initial noise of the
frozen BC policy. Conditioned on the current state, the noise policy and
the resulting action are defined as
\begin{equation}
    \boldsymbol{w}_t
    \sim \pi_{\theta}^{\boldsymbol{w}}(\boldsymbol{w}_t\mid\mathbf{s}_t),
    \qquad
    \mathbf{a}_t
    =\pi_{\mathrm{BC}}(\mathbf{s}_t,\boldsymbol{w}_t).
\end{equation}
Under the original full-dimensional noise parameterization, sampling
$\boldsymbol{w}_t\sim\mathcal{N}(0,\mathbf{I})$ recovers the original BC behavior.
Learning $\pi_{\theta}^{\boldsymbol{w}}$ instead changes the noise distribution and thereby
steers the action generated by $\pi_{\mathrm{BC}}$ without updating the pretrained
policy. The objective is
$
    \max_{\theta}\;
    \mathbb{E}_{\pi_{\theta}^{W}}
    \sum_{t=0}^{T}\gamma^t
    R(\mathbf{s}_t,\mathbf{a}_t)
.
$
Tactile feedback can be used by $\pi_{\theta}^{\boldsymbol{w}}$ to adapt
the frozen BC policy toward successful and contact-safe behaviors.

\section{Methodology}

\methodname combines tactile-conditioned DSRL with physics-guided force
reasoning (PFR) to adapt the behavior of a frozen BC policy. DSRL refines
position alignment and contact timing, while PFR uses physical priors encoded
in a VLM to update a task-specific contact-force goal $g_t$ based on the
interaction history $\boldsymbol{h}_t$ after each episode. 
At each planning
step, the actor $\pi_{\theta}^{\boldsymbol{w}}$ samples latent noise
$\boldsymbol{w}_t$ conditioned on $o_t$, including visual, proprioceptive, and tactile information.
The frozen BC policy maps the noise tensor and
its observations to an action chunk. During contact stage, a tactile-gated low-level force controller adjusts the nominal action $\mathbf{a}_t$ according to
$g_t$ to produce the executed action $\boldsymbol{u}_t$.

The executed action $\boldsymbol{u}_t$ is jointly determined by the RL policy
$\pi_{\theta}^{\boldsymbol{w}}$ and the contact-force goal inferred by PFR.
We treat the original environment and the low-level force controller as a
single effective environment under partial observability. The force goal
$g_t$ is fixed within each episode ($g_t=g_e$) and updated by PFR between
episodes, providing contextual adaptation. We use
$\boldsymbol{z}_t=(o_t,g_t)$ as a context-augmented observation for the
critic $Q(\boldsymbol{z}_t,\boldsymbol{w}_t)$, while the actor remains
conditioned on $o_t$.
We thus model the interaction as a partially observable Markov decision
process (POMDP) and approximate action values using the current
context-augmented observation $\boldsymbol{z}_t$.

\subsection{Real-World Tactile-Informed Latent RL Actor}

\paragraph{VLA and VTLA base policies.}
We initialize the behavior-cloning policy from GR00T N1.7~\cite{gr00tn1_2025}
and fine-tune it on task demonstrations. Each observation contains an image
from the wrist-mounted camera $\mathbf{I}^{\mathrm{wrist}}_t$ and a third-person image
$\mathbf{I}^{\mathrm{ext}}_t$, both resized to $224\times224$ pixels. We
represent the end-effector pose by its Cartesian position and a continuous 6D
orientation representation formed by the first two rows of its rotation
matrix. Together with the current gripper width, the proprioceptive vector is
defined as
\begin{equation}
    \mathbf{p}_t =
    \left[
        (\mathbf{x}^{\mathrm{ee}}_t)^\top,
        \operatorname{vec}(\mathbf{R}^{\mathrm{ee}}_{t,1:2,:})^\top,
        d_t
    \right]^\top \in \mathbb{R}^{10}.
\end{equation}
The VLA policy is conditioned on the two images, $\mathbf{p}_t$, and the
language instruction. For the VTLA policy, we additionally concatenate the 6D
force--torque measurements from the left and right fingertips,
$\mathbf{f}^{L}_t,\mathbf{f}^{R}_t\in\mathbb{R}^{6}$, with $\mathbf{p}_t$.
Thus, the proprioceptive--tactile state has 22 dimensions.

At each planning step $t$, the base policy predicts an action chunk spanning
time steps $t$ through $t+H-1$,
\begin{equation}
    \begin{aligned}
        \mathbf{A}_t
        &= [\mathbf{a}_t,\ldots,\mathbf{a}_{t+H-1}]^\top
        \in \mathbb{R}^{H\times 10}, \\
        \mathbf{a}_{\tau}
        &= \left[
            (\hat{\mathbf{x}}^{\mathrm{ee}}_{\tau})^\top,
            \operatorname{vec}(\hat{\mathbf{R}}^{\mathrm{ee}}_{\tau,1:2,:})^\top,
            \hat d_{\tau}
        \right]^\top
    \end{aligned}
\end{equation}
Each row contains the target position, target orientation, and target gripper
width. We train the policy with an action horizon $H=32$. During deployment, we use receding-horizon control and execute the
first $K=10$ actions before acquiring a new observation and replanning.

\paragraph{Tactile-informed latent actor}
Directly processing raw images in the online RL actor would be computationally
expensive. We therefore reuse visual tokens from the frozen VLM backbone and
pool them into a 2048-dimensional feature $\bar{\mathbf{v}}_t$. For the actor,
$o_t$ is represented by this visual feature, the proprioceptive vector
$\mathbf{p}_t$, and the two fingertip wrenches $\mathbf{f}^{L}_t$ and
$\mathbf{f}^{R}_t$. After modality-specific linear projections, a Transformer
fuses the visual, proprioceptive, and tactile features. Its output head
parameterizes a 10-dimensional latent-noise distribution and samples
$\boldsymbol{w}_t\in\mathbb{R}^{10}$.
GR00T expects a latent-noise tensor in
$\mathbb{R}^{H\times132}$ for each $H$-step action chunk. We repeat the same
10-dimensional actor output across all $H$ chunk positions, producing an
intermediate tensor in $\mathbb{R}^{H\times10}$. At each position, these ten
values occupy the first ten channels. The remaining 122 channels retain their
corresponding entries from a sample drawn once from a standard Gaussian
distribution and held fixed. We include this fixed input expansion in the notation
$\pi_{\mathrm{BC}}(o_t,\boldsymbol{w}_t)$ for the frozen policy's output.
The critics take $\boldsymbol{w}_t$ directly as their action input, so both
actor and critic optimization remain in the 10-dimensional noise space.

\paragraph{Actor--critic optimization.}
Following DSRL-SAC~\cite{Wagenmaker2025SteeringYD}, we treat each executed action
chunk as one RL transition. At chunk index $n$, the replay buffer stores
$(\boldsymbol{z}_n,\boldsymbol{w}_n,r_n,
\boldsymbol{z}_{n+1},\delta_n)$, where $r_n$ is the reward accumulated over the
executed chunk and $\delta_n$ is the terminal flag. We use a standard critic ensemble
$\{Q_{\eta_i}\}_{i\in\mathcal{I}}$ and its corresponding target networks
$\{Q_{\bar\eta_i}\}_{i\in\mathcal{I}}$. For a sampled transition, we draw
$\boldsymbol{w}_{n+1}\sim
\pi_{\theta}^{\boldsymbol{w}}(\cdot\mid o_{n+1})$ and form the chunk-level
temporal-difference target without entropy backup
\begin{equation}
    y_n = r_n + \gamma(1-\delta_n)
    \min_{i\in\mathcal{I}}
    Q_{\bar\eta_i}(\boldsymbol{z}_{n+1},
    \boldsymbol{w}_{n+1}).
\end{equation}
where $\gamma$ is the discount factor between successive action chunks.
The critics are updated by minimizing
\begin{equation}
    \mathcal{L}_{Q}(\{\eta_i\}_{i\in\mathcal{I}})
    =\frac{1}{|\mathcal{I}|}\sum_{i\in\mathcal{I}}
    \mathbb{E}_{\mathcal{B}}
    \left[
        \left(
        Q_{\eta_i}(\boldsymbol{z}_n,
        \boldsymbol{w}_n)-y_n
        \right)^2
    \right].
\end{equation}
We then update the actor through reparameterized samples of
$\boldsymbol{w}_n$ while keeping the critics fixed. The actor objective is
\begin{equation}
    \begin{aligned}
        \mathcal{L}_{\pi}(\theta)
        =\mathbb{E}_{\substack{\boldsymbol{z}_n\sim\mathcal{B},\\
            \boldsymbol{w}_n\sim\pi_{\theta}^{\boldsymbol{w}}}}
        \Big[&\alpha\log
        \pi_{\theta}^{\boldsymbol{w}}(\boldsymbol{w}_n\mid o_n)\\
        &-\min_{i\in\mathcal{I}}
        Q_{\eta_i}\!\left(
            \boldsymbol{z}_n,\boldsymbol{w}_n
        \right)\Big],
    \end{aligned}
\end{equation}
where $\alpha$ controls the exploration entropy and is adjusted toward a target
entropy during training.

\subsection{PFR-Guided Contact-Force Control}

We combine task-specific contact-force bounds inferred by the PFR module
with low-level hybrid force--position control. Different
manipulation tasks require different forms of force regulation. Fragile-object
grasping requires sufficient but non-destructive grasp force, whereas
whiteboard wiping requires an appropriate normal force. Fragile fruit harvesting, like raspberry, requires both a gentle grasp and sufficient pulling force to detach the fruit
from its stem. This combination provides task-specific force feedback during
contact.

We formulate force regulation over contact phases involving gripper--object
interaction, object--environment interaction, or both. Let
$\mathbf{f}_t=[\mathbf{F}_t^\top,\boldsymbol{\tau}_t^\top]^\top\in
\mathbb{R}^{6}$ denote the object--environment wrench in the end-effector frame,
comprising 3D force and 3D torque. When gripper--object interaction must be constrained, such as
to avoid crushing a fragile object, we append the scalar grasp force
$f_t^{\mathrm{g}}$ to obtain
$
    \tilde{\mathbf{f}}_t
    =[(\mathbf{f}_t)^\top,f_t^{\mathrm{g}}]^\top
    \in\mathbb{R}^{7}.
$
We project this vector onto the task-relevant force-control axes as
\begin{equation}
    \mathbf{f}'_t=\mathbf{A}\bar{\mathbf{f}}_t,
    \ \mathbf{A}\in\mathbb{R}^{m\times D},
\end{equation}
where each row of $\mathbf{A}$ defines one regulated force or torque channel and
$m$ is the number of regulated channels. During contact, the projected wrench must
satisfy the component-wise constraint
\begin{equation}
    \mathbf{f}_{\min}\preceq\mathbf{f}'_t\preceq\mathbf{f}_{\max}.
\end{equation}
We initialize $\mathbf{A}$ before RL training using VLM estimates and manual
correction. The PFR module infers
the task-conditioned bounds $\mathbf{f}_{\min}$ and $\mathbf{f}_{\max}$. The
hybrid force--position controller then regulates the selected axes within these
bounds during task execution.

\subsubsection{Force-Control Axis Estimation }

When object--environment force regulation is required, we constrain interaction
along a single task-relevant force or torque axis. Gripper--object interaction
is regulated through the scalar grasp force $f_t^{\mathrm{g}}$ when required.
Thus, $m=1$ for either object--environment regulation alone or grasp-force
regulation alone, and $m=2$ when both channels are regulated.
Before RL training, we use the VLM to estimate the task-relevant principal
force-control axis in the robot's end-effector coordinate frame. At this
stage, we manually review the estimate and correct any erroneous
axis. The resulting force or torque axis defines the corresponding row of
$\mathbf{A}$. When grasp-force regulation is required, $\mathbf{A}$ includes a
row selecting $f_t^{\mathrm{g}}$.

\subsubsection{PFR Module}

After each episode, PFR queries the VLM with the final contact-phase image,
the measured force trajectory, and the force bounds used during execution.
The VLM assesses whether the interaction force was insufficient, excessive,
or adequate, while distinguishing force-related failures from positioning,
trajectory, or force-tracking errors. These diagnoses provide evidence for
estimating the task-specific force bounds.
The diagnosis determines whether and how the force bounds are updated. An
insufficient-force diagnosis shifts the corresponding interval toward a larger
force, whereas an excessive-force diagnosis shifts it toward a smaller force.
When the evidence is indeterminate, the current estimate is retained. 
Likewise, if the failure is attributed to positioning or trajectory execution, the force bounds remain unchanged. 
This outcome-conditioned update has a direct task-dependent interpretation. If
visible marks remain on a whiteboard after wiping, PFR increases the
admissible normal-force range. If a fragile fruit is crushed during harvesting,
it reduces the grasp-force range; if the fruit remains attached to the stem, it
instead increases the pulling-force range. 

To make PFR decision cumulative
rather than myopic, we compress the complete interaction history into the VLM
context. The history records the force bounds of each episode, whether
the episode succeeded or failed, the diagnosed failure cause, and the resulting
update. Consequently, each new estimate is conditioned on both the latest
multimodal evidence and the outcomes of all preceding interactions.

\subsubsection{Stage-aware Force Control}
\begin{figure}[t]
  \centering
  \includegraphics[width=\columnwidth,draft=false]
    {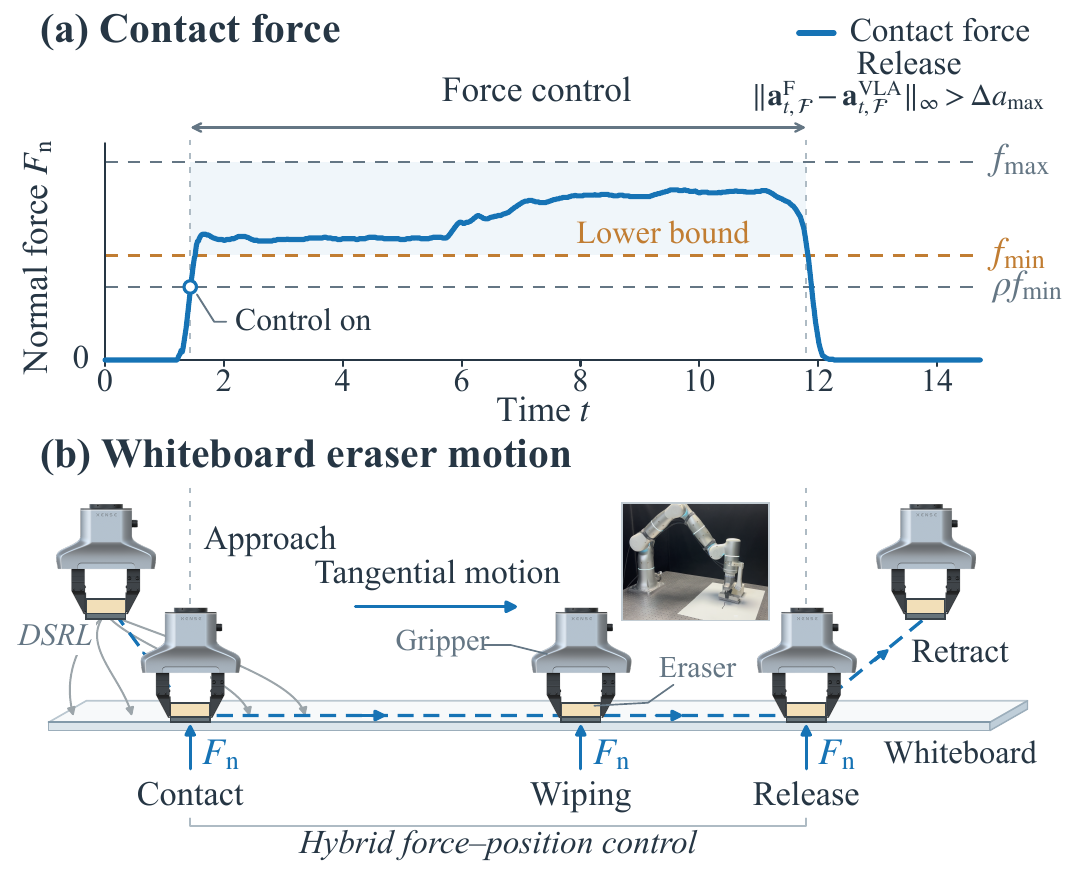}
  \caption{Stage-aware force control during whiteboard wiping.}
  \label{fig:stage_aware_force_control}
\end{figure}

Fig.~\ref{fig:stage_aware_force_control} illustrates the framework of
contact-stage transitions using the whiteboard wiping task.
The low-level force controller runs at $50\,\mathrm{Hz}$, while the policy
runs at $30\,\mathrm{Hz}$.
Force regulation is activated when the manipulation enters a
contact stage, thereby avoiding interference with the policy during
free-space motion. For tasks with a single contact stage, let
$c_t^{\mathrm{bi}}\in\{0,1\}$ indicate that a stable bilateral grasp has been
established, and let $f'_{t,k}$ be the interaction force along the principal
control axis $k$. The contact-stage indicator $\chi_t$ is activated when
both the grasp and load conditions are satisfied,
\begin{equation}
    \chi_t:0\rightarrow1
    \quad\text{if}\quad
    c_t^{\mathrm{bi}}=1
    \ \land
    \lvert f'_{t,k}\rvert\geq \rho f_{\min,k},
    \ \rho=0.7.
\end{equation}
Once activated, an admittance controller modifies the action coordinates
selected by $\mathbf{A}$, while the remaining coordinates continue to follow
the VLA action.
For each regulated channel $k$, the force or torque reference is set to the
midpoint of the PFR-inferred interval,
$f^{\mathrm{ref}}_k=(f_{\min,k}+f_{\max,k})/2$.

We also use the VLA output to determine when force control should be released.
Let $\mathbf{a}^{\mathrm{VLA}}_{t,\mathcal{F}}$ and
$\mathbf{a}^{\mathrm{F}}_{t,\mathcal{F}}$ denote, respectively, the VLA command
and the force-controller command on the force-controlled coordinates
$\mathcal{F}$. Their disagreement is the correction required to preserve force
control. If this correction exceeds the controller's admissible range,
\begin{equation}
    \left\|
    \mathbf{a}^{\mathrm{F}}_{t,\mathcal{F}}-
    \mathbf{a}^{\mathrm{VLA}}_{t,\mathcal{F}}
    \right\|_{\infty}>\Delta a_{\max},
\end{equation}
we set $\chi_{t+1}=0$ and return control of these coordinates to the VLA.
This rule treats a sufficiently large policy command as evidence that the
current contact stage has ended, rather than forcing the controller to oppose
the intended task motion.

During an active contact stage, the low-level controller regulates the selected
axes within the force bounds inferred by PFR. If the measured force reaches the
smaller of $f_{\max,k}$ and the hard safety limit, the controller issues an
unloading command. The executed hybrid action is
\begin{equation}
    \mathbf{a}^{\mathrm{exec}}_t
    =\Psi (\mathbf{a}^{\mathrm{VLA}}_t,\mathbf{a}^{\mathrm{F}}_t,\mathbf{A},\chi_t),
\end{equation}
where $\Psi$ decomposes motion in task space into the force-controlled subspace
specified by $\mathbf{A}$ and its complement. When $\chi_t=1$, motion along the
selected axes follows $\mathbf{a}^{\mathrm{F}}_t$, while motion in the
complementary directions follows $\mathbf{a}^{\mathrm{VLA}}_t$. When $\chi_t=0$,
$\Psi$ returns $\mathbf{a}^{\mathrm{VLA}}_t$.

For tasks involving multiple contact stages, the VLA policy can be extended
to explicitly predict the current contact stage from its observations.
This prediction allows the controller to select the appropriate force-control
axes and PFR-inferred force bounds, enabling finer force regulation at each
stage. For example, during fruit
harvesting, the controller can regulate grasp force when establishing
gripper--fruit contact and additionally regulate pulling force during
fruit--stem detachment, while maintaining a gentle grasp.

\subsection{The Complete System of \methodname}

Algorithm~\ref{alg:rapid-tactile} summarizes the complete pipeline, including
force-axis initialization, online latent RL, stage-aware force
arbitration, and episode-level force adaptation through PFR.

\begin{algorithm}[t]
\caption{\methodname}
\label{alg:rapid-tactile}
\small
\begin{algorithmic}[1]
\Require Demonstrations $\mathcal{D}$; pretrained VLA $\pi_{\mathrm{BC}}$;
VLM $\mathcal{V}$
\State Estimate task-relevant control axes in the end-effector frame
\State Manually review and correct the axes; construct $\mathbf{A}$ from the
selected control channels
\State $[\mathbf{f}_{\min},\mathbf{f}_{\max}]
\gets\Call{PFRInferBounds}{\mathcal{V},\text{task metadata}}$
\State Initialize $\pi_{\theta}^{\boldsymbol{w}}$, critics, 
replay buffer $\mathcal{B}$, and history $\mathcal{H}_{\mathrm{epi}}$
\For{$e=1,\ldots,E$}
    \State Reset the environment; $t\gets0$; $\mathcal{T}_e\gets\varnothing$
    \While{the episode is not terminated}
        \State Observe $o_t$ and tactile wrench $\bar{\mathbf{f}}_t$;
        form $\boldsymbol{z}_t=(o_t,g_e)$
        \State Sample $\boldsymbol{w}_t\sim
        \pi_{\theta}^{\boldsymbol{w}}(\cdot\mid o_t)$
        \State $\mathbf{A}^{\mathrm{VLA}}_t\gets
        \pi_{\mathrm{BC}}(o_t,\boldsymbol{w}_t)$
        \For{$j=t,\ldots,t+K-1$}
            \State $\mathbf{a}^{\mathrm{VLA}}_j\gets
            \mathbf{A}^{\mathrm{VLA}}_t[j-t]$
            \State $\mathbf{f}'_j\gets\mathbf{A}\bar{\mathbf{f}}_j$
            \State $(\chi_j,k_j)\gets
            \Call{ContactStage}{\mathcal{T}_e,\mathbf{f}'_j,
            \mathbf{a}^{\mathrm{VLA}}_j}$
            \If{$\chi_j=1$}
                \State $\mathbf{u}_j\gets\Call{ForceControl}{
                \mathbf{a}^{\mathrm{VLA}}_j,\mathbf{f}'_j,k_j}$
            \Else
                \State $\mathbf{u}_j\gets\mathbf{a}^{\mathrm{VLA}}_j$
            \EndIf
            \State Execute $\mathbf{u}_j$ and append the transition to
            $\mathcal{T}_e$
        \EndFor
        \State Store the chunk transition with $\boldsymbol{w}_t$ in $\mathcal{B}$
        \State Update the critics, latent actor, entropy coefficient, and
        target critics from $\mathcal{B}$
        \State $t\gets t+K$
    \EndWhile
    \State $\Call{PFRUpdateBounds}{\mathcal{V},\mathcal{T}_e,
    \mathcal{H}_{\mathrm{epi}}}$
\EndFor
\end{algorithmic}
\end{algorithm}

%% file: sections/exp.tex
\section{Experiments}
\label{sec:experiments}
\begin{table*}[t]
  \caption{Online adaptation and final-policy performance comparison.
  ``/'' denotes an unreached $E_{80}$.}
  \label{tab:policy_comparison}
  \centering
  \renewcommand{\arraystretch}{1.2}
  \setlength{\tabcolsep}{2pt}
  \resizebox{\textwidth}{!}{%
  \begin{tabular}{l c c c c c c c c c c c c}
    \toprule
    \multirow{2}{*}{\textbf{Method}}
      & \multicolumn{4}{c}{\textbf{\textit{Thin Sheet Transfer}}}
      & \multicolumn{4}{c}{\textbf{\textit{Bottle Cap Twisting}}}
      & \multicolumn{4}{c}{\textbf{\textit{Fragile Fruit Picking}}} \\
    \cmidrule(lr){2-5}\cmidrule(lr){6-9}\cmidrule(lr){10-13}
      & \shortstack{SR (\%)\\Mean $\pm$ Std. $\uparrow$}
      & \shortstack{nAUC (\%)\\$\uparrow$}
      & \shortstack{$E_{80}$\\$\downarrow$}
      & \shortstack{Force failure (\%)\\$\downarrow$}
      & \shortstack{SR (\%)\\Mean $\pm$ Std. $\uparrow$}
      & \shortstack{nAUC (\%)\\$\uparrow$}
      & \shortstack{$E_{80}$\\$\downarrow$}
      & \shortstack{Force failure (\%)\\$\downarrow$}
      & \shortstack{SR (\%)\\Mean $\pm$ Std. $\uparrow$}
      & \shortstack{nAUC (\%)\\$\uparrow$}
      & \shortstack{$E_{80}$\\$\downarrow$}
      & \shortstack{Force failure (\%)\\$\downarrow$} \\
    \midrule
    Base Policy
      & $73.7 \pm 0.6$ & $73.7$ & / & $5.0$
      & $43.0 \pm 1.0$ & $43.0$ & / & $46.0$
      & $23.7 \pm 1.2$ & 23.7 & / & $52.3$ \\
    Residual RL \cite{Yu2026OmniTacTunePR}
      & $75.7 \pm 0.6$ & $76.0$ & 12 & $6.7$
      & $68.0 \pm 20.3$ & $68.0$ & $47$ & $20.3$
      & $37.3 \pm 2.1$ & $23.2$ & / & $44.7$ \\
    AWR \cite{peng2019advantage}
      & $67.7 \pm 2.5$ & $68.2$ & $37$ & $7.7$
      & $21.0 \pm 13.1$ & $20.3$ & / & $61.7$
      & $15.7 \pm 1.2$ & $16.4$ & / & $55.7$ \\
    DPPO \cite{ren2025diffusion}
      & $4.7 \pm 3.5$ & $2.5$ & / & $5.7$
      & $32.7 \pm 1.5$ & $31.7$ & / & $55.3$
      & $22.3 \pm 2.3$ & $16.3$ & / & $48.0$ \\
    DSRL \cite{Wagenmaker2025SteeringYD}
      & $82.3 \pm 4.5$ & $72.2$ & $29$ & $9.7$
      & $57.7 \pm 1.5$ & $57.9$ & / & $31.3$
      & $40.3 \pm 4.2$ & $25.3$ & / & $36.3$ \\
    \midrule
    \textbf{\methodname}
      & $\mathbf{94.7} \pm 1.5$ & $\mathbf{94.7}$ & $\mathbf{10}$ & $\mathbf{2.0}$
      & $\mathbf{87.7} \pm 1.5$ & $\mathbf{90.1}$ & $\mathbf{22}$ & $\mathbf{10.3}$
      & $\mathbf{77.7} \pm 1.2$ & $\mathbf{73.4}$ & $\mathbf{25}$ & $\mathbf{3.3}$ \\
    \bottomrule
  \end{tabular}
  }
\end{table*}
We organize the experiments around three questions:
\textbf{Q1: Adaptation efficiency.} Does \textbf{\methodname} enable more sample-efficient online adaptation than existing policy
fine-tuning methods?
\textbf{Q2: PFR module.} Can the PFR module in \textbf{\methodname} infer task-appropriate contact-force bounds, and how robust are these predictions?
\textbf{Q3: Real-world deployment.} Can \textbf{\methodname} be deployed effectively on physical robotic systems?

\subsection{Experimental Setup}
\subsubsection{Evaluation Tasks}

We consider three simulation tasks: Thin Sheet Transfer, Bottle Cap Twisting,
and Fragile Fruit Picking. In Thin Sheet Transfer, the robot must transport a
fragile thin sheet to the target and
regulating its grasp force to avoid crushing it. The object position is
randomized within a $0.2\,\mathrm{m}\times0.2\,\mathrm{m}$ region, and the
maximum allowable interaction force is $1.1$--$1.6\,\mathrm{N}$. 
In Bottle Cap Twisting, the robot must tighten a bottle cap and establish an
appropriate normal force between the cap and the bottle after tightening. This
task evaluates whether the PFR module can infer an appropriate
interaction torque. The object position is randomized within a
$0.1\,\mathrm{m}\times0.1\,\mathrm{m}$ region, and the target torque range for
successful completion is $0.35$--$0.50\,\mathrm{N\,m}$. 
In Fragile Fruit Picking, the robot must detach a raspberry from its vine. It
must apply a gentle grasp to avoid crushing the fruit while exerting sufficient
pulling force for detachment. The object position is randomized by up to
$\pm[8,10,12]\,\mathrm{mm}$ along the $x$-, $y$-, and $z$-axes,
respectively. The crushing-force threshold is $3.2\,\mathrm{N}$, and the
minimum force required for detachment is $0.8\,\mathrm{N}$.
The preset open-loop pulling offset is zero.

For real-world evaluation, we consider two tasks: Whiteboard Erasing and
Pipette Liquid Aspiration. In Whiteboard Erasing, the robot must learn to apply
an appropriate contact force between the eraser and the whiteboard to remove
the marker traces. In Pipette Liquid Aspiration, the robot must regulate the
fingertip contact force on the pipette: the force must be sufficient to lift
the pipette while remaining low enough to prevent the liquid inside from being
expelled.

For each task, we use 50 offline demonstration trajectories.
Base policies are obtained by fine-tuning GR00T N1.7~\cite{gr00tn1_2025}
on the corresponding offline demonstrations. 
For Thin Sheet Transfer and real-world tasks, the
base policies do not use tactile features. For Bottle Cap Twisting and Fragile
Fruit Picking, the state additionally includes the 6D force--torque
measurements from both fingertips. All simulation experiments are conducted in
NVIDIA Isaac Sim.

\subsubsection{Training and Evaluation Detail}

The GR00T-based policies are trained on an NVIDIA H100 GPU for a fixed 30 epochs, and
the checkpoint from the final epoch is used as the base policy. Training takes
approximately five hours per task. For the baseline comparisons, simulation-based
RL training is performed on an NVIDIA GeForce RTX 5090 GPU, with GPT-5.6 Sol configured at
extra-high reasoning effort as the VLM backbone of PFR. For the real-world experiments, policy
inference and online RL training are also performed on an NVIDIA GeForce RTX
5090 GPU.

\subsubsection{Baselines}
We compare \textbf{\methodname} with a \textbf{frozen base policy} and four representative
online policy-adaptation methods. \textbf{Residual RL}~\cite{ankile2025residual,Yu2026OmniTacTunePR}
uses a lightweight residual network to correct the actions generated by the
frozen base policy, whereas \textbf{DSRL}~\cite{Wagenmaker2025SteeringYD} learns a
tactile-conditioned actor in the policy's latent noise space. \textbf{AWR}~\cite{peng2019advantage}
updates the policy through advantage-weighted regression, and \textbf{DPPO}~\cite{ren2025diffusion}
uses a PPO-style objective to update the diffusion policy. 
\textbf{\methodname}, \textbf{Residual RL}, and \textbf{DSRL}
combine a tactile-conditioned actor.
For all RL algorithms, we use a sparse terminal reward of $+10$ for a
successful episode and $-10$ for a failed episode, with a discount factor of
$\gamma=0.9$. 
In simulation, success or failure is determined automatically
by the simulator, whereas real-world outcomes are labeled manually by a human
operator.

\subsection{Comparison with Baselines (\textbf{Q1})}

\begin{figure*}[t]
  \centering
  \includegraphics[width=\textwidth]{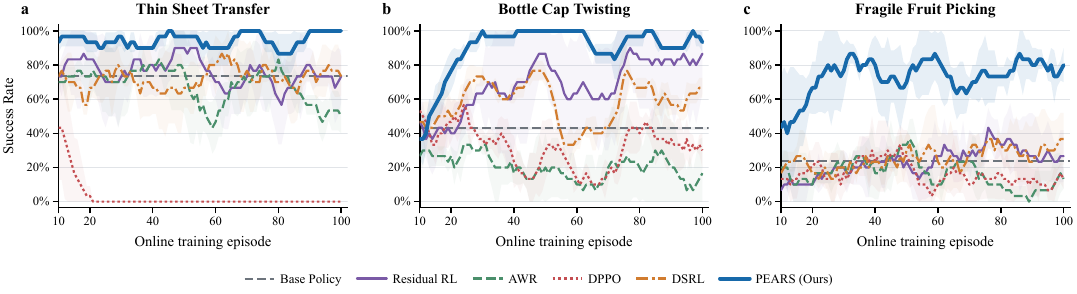}
  \caption{Online adaptation success across three simulation tasks. Curves show mean trailing-10 success rates across three
  100-episode runs, using complete windows ending at episodes 10--100.
  Shaded bands denote mean $\pm$ one sample standard deviation (SD)}
  \label{fig:online_success}
\end{figure*}

Each adaptive method used three 100-episode online runs, followed by
100 test trials per run. Final-policy success rate (SR) is reported as
mean $\pm$ sample SD. Let $\mathrm{SR}_e\in[0,1]$ denote trailing-10
online success averaged across runs (Fig.~\ref{fig:online_success}). Normalized
area under the learning curve (nAUC) is the mean of $\mathrm{SR}_e$ for
$e=10,\ldots,100$, expressed as a percentage; $E_{80}$ is the first
episode with $\mathrm{SR}_e\geq0.8$ (``/'' if unreached).
nAUC measures overall online performance, while a smaller $E_{80}$
indicates faster attainment of a high success rate.
Force-related failure rate is the percentage of all episodes ending in
task-specific force or contact failures, pooled across runs.

\methodname achieved the highest SR on Thin Sheet Transfer, Bottle Cap
Twisting, and Fragile Fruit Picking: $94.7\%$, $87.7\%$, and $77.7\%$
(Table~\ref{tab:policy_comparison}). These exceed the strongest per-task
baselines by 12.4, 19.7, and 37.4 percentage points. It also led in nAUC
($94.7\%$, $90.1\%$, and $73.4\%$), indicating higher success throughout
online adaptation. The largest SR gain occurs on Fragile Fruit Picking,
which requires coordinated grasping and pulling without damaging the fruit.

\methodname reached $E_{80}$ at episodes 10, 22, and 25. On Thin Sheet
Transfer, episode 10 is the earliest evaluable window, compared with
12 for the fastest baseline, Residual RL. On Bottle Cap Twisting,
Residual RL required 47 episodes, so \methodname reduced interactions
to the threshold by $53.2\%$. No baseline reached $80\%$ on Fragile
Fruit Picking within 100 episodes. Higher success was accompanied by
the lowest force-related failure rates ($2.0\%$, $10.3\%$, and $3.3\%$),
compared with $9.7\%$, $20.3\%$, and $36.3\%$ for the strongest SR baselines.
The mean waiting time for a VLM response was 26.8~s, which is acceptable
because this wait can overlap with other steps, such as environment resets.

\subsection{Evaluation of the PFR Module (\textbf{Q2})}

\begin{figure}[t]
  \centering
  \includegraphics[width=\columnwidth]{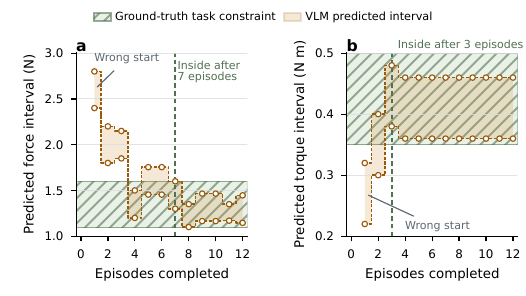}
  \caption{Example of force and torque intervals predicted by PFR during online interaction. a) Thin Sheet Transfer. b) Bottle Cap Twisting. }
  \label{fig:force_interval_adaptation}
\end{figure}

Fig.~\ref{fig:force_interval_adaptation} shows how PFR corrects
initially inaccurate force and torque intervals in two tasks.
Over the first 12 episodes, predicted force and torque intervals
stabilize within ground-truth ranges, illustrating refinement of
initial estimates from interaction outcomes.

Table~\ref{tab:vlm_dsrl_ablation} compares five variants.
\textit{DSRL only} omits PFR and external force control;
\textit{w/o DSRL} combines PFR and force control with the frozen base
policy. \textit{Fixed Range} and \textit{Rand Search} retain DSRL and
force control, respectively holding the first PFR-inferred intervals
fixed or randomly searching intervals online.
For \textit{Rand Search}, the checkpoint with the highest mean online success is used for evaluation.
\textit{Qwen3} replaces GPT-5.6 Sol with
Qwen3-VL-2B~\cite{bai2025qwen3} in PFR.

\begin{table}[t]
  \caption{Ablation of PFR, DSRL, force-range selection, and the VLM
  backbone used in PFR across three simulation tasks.}
  \label{tab:vlm_dsrl_ablation}
  \centering
  \setlength{\tabcolsep}{2pt}
  \renewcommand{\arraystretch}{1.1}
  \resizebox{\columnwidth}{!}{%
  \begin{tabular}{@{}l c c c c c c@{}}
    \toprule
    SR (\%) & DSRL only & w/o DSRL & Fixed Range & Rand Search & Qwen3
      & \textbf{PEARS} \\
    \midrule
    Sheet & $82.3\,{\scriptstyle\pm 4.5}$ & $77.0\,{\scriptstyle\pm 1.0}$ & $84.7\,{\scriptstyle\pm 3.8}$ & $88.3\,{\scriptstyle\pm 6.0}$ & $93.0\,{\scriptstyle\pm 1.7}$ & $\mathbf{94.7}\,{\scriptstyle\pm 1.5}$ \\
    Cap & $57.7\,{\scriptstyle\pm 1.5}$ & $84.7\,{\scriptstyle\pm 1.5}$ & $62.3\,{\scriptstyle\pm 2.5}$ & $77.0\,{\scriptstyle\pm 5.6}$ & $86.7\,{\scriptstyle\pm 0.6}$ & $\mathbf{87.7}\,{\scriptstyle\pm 1.5}$ \\
    Fruit & $40.3\,{\scriptstyle\pm 4.2}$ & $61.7\,{\scriptstyle\pm 2.1}$ & $44.0\,{\scriptstyle\pm 2.6}$ & $40.3\,{\scriptstyle\pm 2.5}$ & $67.0\,{\scriptstyle\pm 2.6}$ & $\mathbf{77.7}\,{\scriptstyle\pm 1.2}$ \\
    \bottomrule
  \end{tabular}%
  }
\end{table}

On Thin Sheet Transfer, DSRL alone achieved $82.3\%$ SR, compared with
$77.0\%$ for PFR-guided force control without DSRL. Combining both
increased SR to $94.7\%$, versus $84.7\%$ with fixed intervals.
On Bottle Cap Twisting, PFR-guided force control without DSRL retained
$84.7\%$ SR, whereas DSRL alone achieved $57.7\%$ and
\textit{Fixed Range} $62.3\%$. The full method reached $87.7\%$,
indicating a greater benefit from adapting interaction torque.
The different ablation rankings suggest that motion adaptation
contributes more to Thin Sheet Transfer, whereas force control is more
important for Bottle Cap Twisting.
On Fragile Fruit Picking, removing PFR and force control or removing
DSRL reduced SR from $77.7\%$ to $40.3\%$ or $61.7\%$, respectively;
fixed intervals yielded $44.0\%$. Fixed intervals provide only modest
gains over DSRL across tasks, suggesting that online correction is
important when initial bounds are inaccurate. These results support
complementary roles for learned motion adaptation and online
force-interval correction.

\textit{Rand Search} increased SR over DSRL 
on the two single-interval tasks, but remained below \methodname.
Random search can find partially suitable bounds in these tasks.
For example, in one Thin Sheet Transfer run, its interval
$[1.32,1.98]\,\mathrm{N}$ partially overlaps the ground-truth range
$[1.1,1.6]\,\mathrm{N}$. This permits useful forces, while the excessive
upper bound helps explain the remaining performance gap.
On two-stage Fragile Fruit Picking, its $40.3\%$ mean SR matches DSRL,
consistent with the difficulty of randomly identifying suitable
grasping and pulling intervals jointly. A usable grasping interval
alone cannot ensure success if pulling remains ineffective, so useful
updates must account for both contact stages.
The difference is also evident in adaptation speed.
On Thin Sheet Transfer and Bottle Cap Twisting, \methodname reached
$E_{80}=10$ and $22$, versus $55$ and $83$ for \textit{Rand Search}.
Earlier attainment of $80\%$ mean online success suggests that VLM
reasoning provides useful direction for interval updates and improves
online sample efficiency.

Qwen3 achieved SRs of $93.0\%$, $86.7\%$, and $67.0\%$, trailing
\methodname by 1.7, 1.0, and 10.7 percentage points, respectively.
Both VLM backbones improve over DSRL alone across tasks, indicating
that the lightweight model still provides useful force guidance.
The larger gap on Fragile Fruit Picking suggests that reasoning
quality matters more when inferring force bounds with
multiple stages.

\subsection{Real-world Experiment (\textbf{Q3})}

\begin{table}[t]
  \caption{Real-world performance (successful trials / total trials).}
  \label{tab:real_world_comparison}
  \centering
  \begin{tabular*}{\columnwidth}{@{\extracolsep{\fill}}l c c c@{}}
    \toprule
    Task
      & Base Policy
      & DSRL \cite{Wagenmaker2025SteeringYD}
      & \textbf{\methodname} \\
    \midrule
    Whiteboard Erasing & 11 / 20 & 15 / 20 & 19 / 20\\
    Pipette Liquid Aspiration & 8 / 20 & 12 / 20 & 18 / 20\\
    \bottomrule
  \end{tabular*}
\end{table}

For each real-world task, DSRL and \methodname are allocated 40 episodes
of online interaction. We then evaluate each adapted policy and the frozen
base policy over 20 test trials (Table~\ref{tab:real_world_comparison}).
In Whiteboard Erasing, sufficient normal force is needed to remove marker
traces, whereas excessive force can dislodge the eraser from the gripper.
The base policy succeeds in $11/20$ trials; its failures comprise three
wiping misalignments, two contact-timing errors that leave residual marks,
and four cases of insufficient force. DSRL improves success to $15/20$,
with two misalignment failures and three insufficient-force failures.
\methodname reaches $19/20$, with the sole failure caused by misalignment.
In Pipette Liquid Aspiration, any liquid leakage during transfer counts as
a failure. The base policy succeeds in $8/20$ trials, with four initial
grasping misalignments, six cases of leakage during transfer, and two
failures during final dispensing. 
DSRL reaches $12/20$, with two grasping
misalignments, five cases of liquid expulsion during transfer,
and one dispensing failure. \methodname succeeds in $18/20$, with both
failures caused by initial grasping misalignment.

These results suggest that DSRL alone struggles to learn precise force
regulation within 40 episodes, particularly for liquid transfer. The absence
of force-related failures with \methodname in these tests supports the
benefit of using PFR to adapt force bounds from tactile feedback and
enforcing them through low-level force control.

%% file: sections/conclusion.tex
\section{Conclusion \& Future Work}

We presented \methodname, a framework for rapid post-training
of pretrained policies that combines PFR with traditional DSRL. PFR uses
tactile feedback and visual outcomes to diagnose failure causes and adapt
contact-force bounds, which are enforced by a low-level controller, while
DSRL refines the behavior of the frozen base policy. Across three simulation
tasks, \methodname improved success rates by 12.4--37.4 percentage points
over the strongest baseline for each task, with faster online adaptation
and fewer force-related failures. In real-world tasks, \methodname achieved $19/20$ and $18/20$ successful test trials on
Whiteboard Erasing and Pipette Liquid Aspiration, respectively.

Currently, PFR provides corrective guidance only for failures caused by
inappropriate force; other errors still rely on
conventional RL for correction. Future work will explore integrating failure
reasoning more broadly into RL to guide exploration and policy updates
beyond force regulation. This could reduce costly trial-and-error
interaction and improve the efficiency of learning on physical robots.